\documentclass[10pt,twocolumn,letterpaper]{article}

\usepackage{iccv}
\usepackage{times}
\usepackage{epsfig}
\usepackage{graphicx}
\usepackage{amsmath}
\usepackage{amssymb}
\usepackage{booktabs}
\usepackage{float}
\usepackage{bm}
\usepackage{multirow}
\usepackage{ulem}
\usepackage{stmaryrd}
\usepackage{color}
\usepackage[accsupp]{axessibility} 

\usepackage{array}
\usepackage{arydshln}
\usepackage[dvipsnames]{xcolor}
\definecolor{citec}{RGB}{21, 101, 192}
\definecolor{refc}{RGB}{220, 40, 40}
\usepackage[pagebackref=true,breaklinks=true,letterpaper=true,colorlinks,bookmarks=false,citecolor=citec,linkcolor=refc]{hyperref}

\iccvfinalcopy % *** Uncomment this line for the final submission

\def\iccvPaperID{****} % *** Enter the ICCV Paper ID here

\ificcvfinal\fi

\begin{document}

%%%%%%%%% TITLE
\title{Learning Object-Compositional Neural Radiance Field  \\ for Editable Scene Rendering
\vspace{-1em}
}

\author{
Bangbang Yang$^{1}$\\
Han Zhou$^{1}$
\and
Yinda Zhang$^{2}$\\
Hujun Bao$^{1}$
\and
Yinghao Xu$^{3}$\\
Guofeng Zhang$^{1}$
\and
Yijin Li$^{1}$ \\
Zhaopeng Cui$^{1}$\thanks{}
\and
$^{1}$State Key Lab of CAD\&CG, Zhejiang University \quad
$^{2}$Google  \quad
$^{3}$The Chinese University of Hong Kong \\
}

\twocolumn[{%
\vspace{-1.6em}
\maketitle
\begin{figure}[H]
\hsize=\textwidth
\centering
\vspace{-3em}
\includegraphics[width=2.0\linewidth, trim={0 0 0 0}, clip]{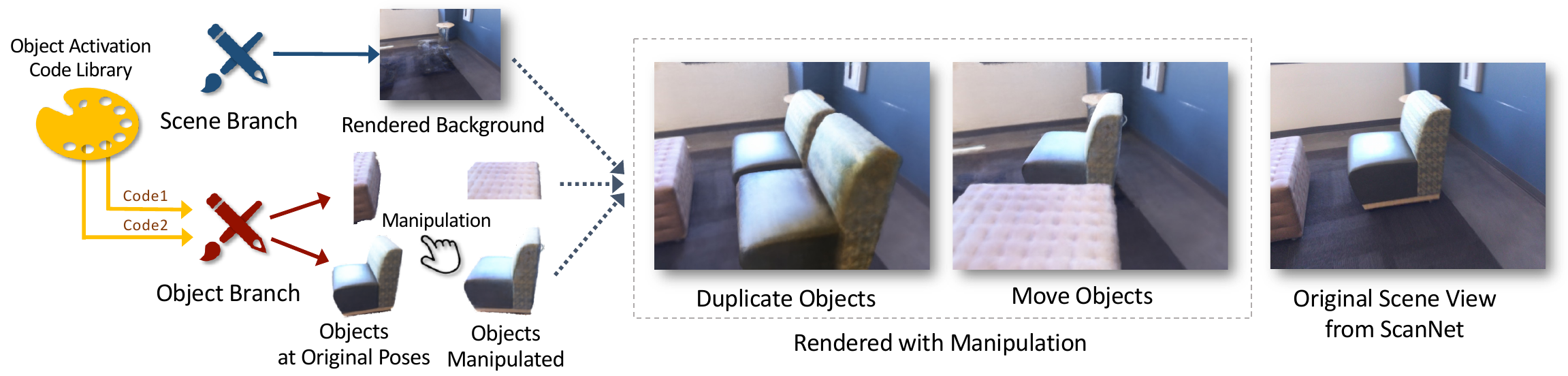}
\caption{
We propose a novel object-compositional neural radiance field that supports editable scene rendering on real-world datasets.
To obtain a view with object manipulation,
we jointly render the transformed objects from the conditioned object branch and the surrounding background from the scene branch.
}
\vspace{-0.0em}
\label{fig:main}
\end{figure}
}]

{
  \renewcommand{\thefootnote}%
    {\fnsymbol{footnote}}
  \footnotetext[1]{Corresponding author}
  \footnotetext{Code is available on the project webpage: \url{https://zju3dv.github.io/object_nerf/}}
}

\ificcvfinal\thispagestyle{empty}\fi
%%%%%%%%% ABSTRACT

% \begin{abstract}
\begin{abstract}
Implicit neural rendering techniques have shown promising results for novel view synthesis.
However, existing methods usually encode the entire scene as a whole, 
which is generally not aware of the object identity and limits the ability to the high-level editing tasks such as moving or adding furniture.
In this paper, we present a novel neural scene rendering system, 
which learns an object-compositional neural radiance field and produces realistic rendering with editing capability for a clustered and real-world scene.
Specifically, we design a novel two-pathway architecture, in which the scene branch encodes the scene geometry and appearance, and the object branch encodes each standalone object conditioned on learnable object activation codes.
To survive the training in heavily cluttered scenes, we propose a scene-guided training strategy to solve the 3D space ambiguity in the occluded regions and learn sharp boundaries for each object.
Extensive experiments demonstrate that our system not only achieves competitive performance for static scene novel-view synthesis, but also produces realistic rendering for object-level editing.

\vspace{-2.0em}
\end{abstract}

%%%%%%%%% BODY TEXT
\section{Introduction}
Virtual tour in a real-world scene is one of the most desired experiences for virtual and augmented reality.
While early works rely on laborious capturing and reconstruction of the physical world, \eg, geometry, texture, material, etc., the emerging neural rendering methods open great opportunities to ease this task by learning directly from a collection of posed images and achieve promising realistic images.
A common follow-up question to ask is: Can we modify the scene, \eg, moving or adding furniture, while still maintaining the realistic rendering capability.

Unfortunately, this is not well-supported by existing neural rendering methods.
Early approaches tend to encode the entire visible scene into a single neural network, such as NeRF~\cite{nerf} and SRN~\cite{SRN}.
While handling small objects perfectly, these models are hard to scale up for large-scale scenes due to the fixed network capacity.
On the other hand, a family of neural rendering approaches utilizes volumetric representation~\cite{nsvf} to densely encode local information at specific locations, which migrates the scalability burden from network parameters to the scene representation and empirically produces better rendering quality.
However, the scene representation and rendering network are in general agnostic to the object identity, which does not support high-level editing tasks such as moving furniture.

In this paper, we propose a neural rendering system that enables scene editing on real-world scenes.
Taking a collection of posed images captured from the real scene and rough 2D instance masks, our model can render the whole scene as it is in reality, as well as with objects manipulated, such as moving, rotating, or duplicating.
Most related to us, OSF~\cite{osf} enables editable scene rendering in a bottom-up fashion by learning one model per-object and then perform joint rendering.
However, their method does not learn the object arrangements in the real world and requires training images captured for each individual object beforehand, which is infeasible to obtain on cluttered scene images and thus only verified on synthetic data.
In contrast, we aim to design a top-down approach that directly learns a unified neural rendering model for the whole scene which respects the object placement as in the captured scene. 
To support object manipulation, we design a novel conditional neural rendering architecture that is able to render each object standalone with everything else removed, which can be further rendered from a novel viewpoint, at a new location, or replicated.
Note that to ensure realistic scene editing, each object has to be rendered with sharp boundaries without background bleeding, which is infeasible to achieve with only a rough 3D rendering mask or a bounding box (see Fig.~\ref{fig:compare_render_obj} for an example).

Indeed, it is non-trivial to learn such an object-compositional neural radiance field for a \textit{clustered} and \textit{real-world} scene even with rough 2D instance masks, mainly due to the 3D space ambiguity in the occluded region.
Intuitively, the network could learn only from the rays casting within the instance mask of a particular object when aiming to render it. 
However, without known geometry, it is impossible to identify if a 3D location belongs to the object but occluded, which is common in a cluttered scene, or not even a part of it, since both cases are not marked by the instance mask.
As a result, the network may overkill part of the object and produce cloudy results.
In order to solve this challenge, we learn an extra compact scene branch, without editable capability, to provide biased sampling distribution along the ray and dense depth online during training, which helps to identify the occluded region where the no gradient is applied instead of being supervised as empty space.
The scene branch also renders the contents that are not labeled by the instance segmentation to provide a seamless whole scene rendering.

In summary, the contributions of this paper are as follows.
Firstly, we propose the first editable neural scene rendering system given a collection of posed images and 2D instance masks, which supports high-quality novel view rendering as well as object manipulation.
Secondly, we design a novel two-pathway architecture to learn object-compositional neural radiance field for a clustered and real-world scene resolving occlusion ambiguity.
Lastly, the experiment and extensive ablation study demonstrate the effectiveness of our system and the design of each component. Our system performs on-par or even better than the SoTA methods in terms of standard novel-view synthesis while maintaining the capability of editable scene rendering with high quality.

\begin{figure*}[!htbp]
    \vspace{-2.5em}
    \centering
    \includegraphics[width=0.9\linewidth, trim={0 0 0 0}, clip]{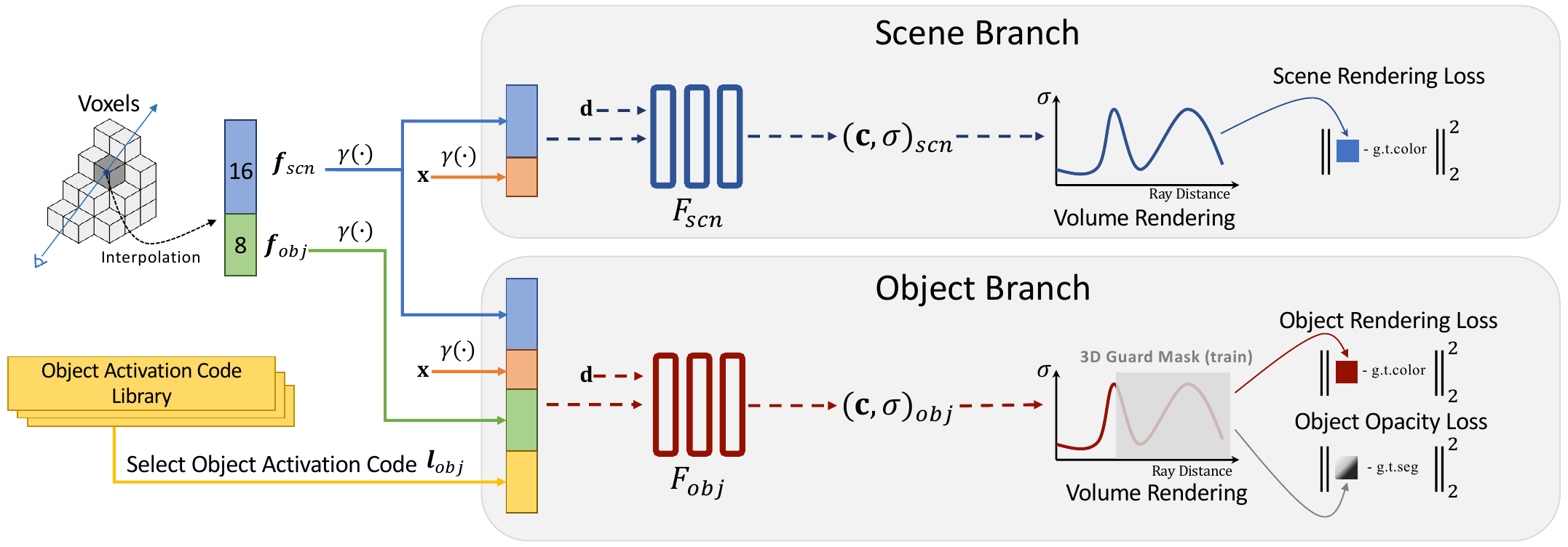}
    \caption{
    We design a two-pathway architecture for object-compositional neural radiance field.
    The scene branch takes the spatial coordinate $\mathbf{x}$, the interpolated scene voxel features $\bm{f}_{scn}$ at $\mathbf{x}$ and the ray direction $\mathbf{d}$ as input,
    and output the color $\mathbf{c}_{scn}$ and opacity $\sigma_{scn}$ of the scene.
    The object branch takes additional object voxel features $\bm{f}_{obj}$ as well a a object activation code $\bm{l}_{obj}$ to condition the output only contains the color $\mathbf{c}_{obj}$ and opacity $\sigma_{obj}$ for a specific object at its original location with everything else removed.
    }
    \label{fig:framework}
    \vspace{-1.5em}
\end{figure*}

\section{Related Work}
\noindent\textbf{Neural Rendering.}
In these works, deep neural networks are employed to learn novel view synthesis from 2D images in various approaches, such as voxels~\cite{sitzmann2019deepvoxels_voxel1, lombardi2019neural_voxel2}, point clouds~\cite{aliev2019neural_point1, pittaluga2019revealing_point2}, textured meshes~\cite{thies2019deferred_mesh1, liu2019neural_mesh2, liao2020towards_mesh3}, multi-plane images~\cite{thies2019deferred_mesh1, zhou2018stereo_multiplane2, npcr}, and implicit functions~\cite{SRN, liu2020dist_implicit2, niemeyer2020differentiable_implicit3}. 
As a pioneer, SRN~\cite{SRN} represents a continuous scene as an opaque surface by implicitly mapping space coordinates to a feature vector with MLPs, and uses a differentiable ray marching algorithm to render 2D feature maps for image generation.
NeRF~\cite{nerf} represent scenes with implicit fields of volume density and view-dependent color, and achieves photo-realistic novel view synthesis results.
To accelerate the rendering speed and enlarge the network capacity, NSVF~\cite{nsvf} proposes a sparse voxel octree variant of NeRF, which implicitly encodes the local properties of a visible scene inside the voxel-bounded representations.
However, as these methods tend to encode the entire scene, it is not trivial to render an individual object once the model has been trained.
On the contrary, our proposed object-compositional neural radiance field naturally supports the standalone object rendering.
On this basis, we realize the novel view synthesis with user-defined object manipulation.

\noindent\textbf{Object-Decomposite Rendering.}
Early approaches adopt the traditional modeling~\cite{mvs1, mvs2, mvs3, mvs5, mvs6, colmap, schoenberger2016mvs} and rendering pipeline~\cite{karsch2011rendering} to support editing and novel view synthesis.
For example, Karsch \etal~\cite{karsch2011rendering} propose to realistically insert synthetic objects into photographs by estimating environment light conditions.
Cossairt \etal~\cite{cossairt2008light} composite real and synthetic objects together with a light field interface, while the object light field is captured with a specific hardware system.
Recently, some works adopt neural implicit representations for object-decomposite rendering.
Guo~\etal~\cite{osf} propose a bottom-up method by learning one scattering field per-object and enables rendering scenes with moving objects and lights, but it needs to train each separate model on images that only contain a single specific object, which is impractical for real-world scenarios.
Ost \etal~\cite{scene-graph-dynamic} propose to use a neural scene graph to decompose dynamic objects in a street view dataset, but their method relies on a dynamic scene, which is not capable of indoor scene scans.
Besides, the latent class encoding restricts each model to represent only one class of objects with similar shapes and canonical coordinates, which limits the applications in general cases where the objects vary and do not share the same shape characteristics.
In contrast, our method does not rely on the canonical coordinates of objects and can also simultaneously learn a compact object-compositional model which enables novel view synthesis with multi-objects manipulation in real-world scene scans.

\section{Method}

\subsection{Overview}

Our framework consists of two pathways: the scene branch and the object branch,
as illustrated in Fig.~\ref{fig:framework}.
The scene branch aims to encode the entire scene geometry and appearance, which renders the surrounding background in editable scene rendering and assists the object branch in identifying the occlusion region.
With 2D instance masks as guidance, the object branch encodes each standalone object conditioned on several learnable object activation codes.
At the rendering stage, when conditioning the scene branch with the object activation code, we can freely render a single object while removing everything else.
It is noteworthy that our framework simultaneously learns to encode multiple objects by assigning a bunch of shuffled object activation codes to the training rays, without the need to train for each object separately.
Since the framework is built upon NeRF, we refer to Mildenhall \etal~\cite{nerf} for the technical background.

\subsection{Framework of Object-Compositional NeRF}
As shown in Fig.~\ref{fig:framework}, our framework adopts two separate branches for scene rendering and object rendering.
We take the advantages both from the voxelized representation~\cite{nsvf} and the coordinate-based positional encoding~\cite{nerf}, and propose a hybrid space embedding as network input.
Practically, for each point $\mathbf{x}$ sampled along the camera ray, we apply positional encoding $\gamma(\cdot)$~\cite{nerf} on both of the scene voxel feature $\bm{f}_{scn}$ interpolated from 8 nearest vertices and space coordinate $\mathbf{x}$ to get the hybrid space embedding.
This hybrid space embedding, along with the embedded directions $\gamma(\bm{d})$, will be fed into the scene branch and the object branch.
By now, the scene branch function $F_{scn}$ can output the opacity $\sigma_{scn}$ and color $\mathbf{c}_{scn}$ of the scene at $\mathbf{x}$.
For the object branch function $F_{obj}$, we additionally add embedded object voxel feature $\gamma(\bm{f}_{obj})$ and object activation code $\bm{l}_{obj}$ to the input, where $\bm{f}_{obj}$ helps to broaden the ability of learning decomposition and is shared by all the objects, and $\bm{l}_{obj}$ identifies feature space for different objects and is possessed by each individual.
Take the object activation code $\bm{l}_{obj}$ as a condition, the object branch precisely outputs color $\mathbf{c}_{obj}$ and opacity $\sigma_{obj}$ for the desired object while everything else remains empty.

\subsection{Object-Compositional Learning}
\label{ssec:object_learn}

\noindent\textbf{Object supervision.}
Ideally, an object radiance field should only be opaque at the area occupied by the object and transparent elsewhere (\ie, zero opacity).
To achieve this goal, we leverage 2D instance segmentation as supervision signals for the object branch.
For brevity, we assume a training process with $K$ annotated objects in a scene, along with a learnable object code library $\mathcal{L}=\{{\bm{l}_{obj}^k}\}$.
For each ray $\bm{r}$ in the batched training set $N_r$, we select one object $k$ as a training target and assign the object activation code ${\bm{l}_{obj}^k}$ to the object branch input.
Then we forward the network and acquire the rendered color $\hat{C}{(\bm{r})^k_{obj}}$, as well as the rendered 2D object opacity $\hat{O}{(\bm{r})_{obj}^k}$ by summing up the product of transmittance $T_i^k$ and alpha value $\alpha_i^k$ of $N$ sampled points along the ray, which follows~\cite{quadrature_rule, nerf} and is defined as \footnote{For brevity, we omit $k$ in $ \hat{C}(\bm{r})_{obj}^k, \hat{O}(\bm{r})_{obj}^k, T_i^k, \alpha_i^k, {\mathbf{c}^k_{obj}}_i, {{\sigma}^k_{obj}}_i$.}:
\begin{equation}
\label{eq:obj}
\vspace{-0.5em}
\begin{split}
    &\hat{C}(\bm{r})_{obj} = \sum_{i=1}^{N}
    T_i \alpha_i {\mathbf{c}_{obj}}_i, \;\;
    {\hat{O}(\bm{r})_{obj}} = \sum_{i=1}^{N}
    T_i \alpha_i, \\ 
    & T_i = \exp{\left(-\sum_{j=1}^{i-1}{\sigma_{obj}}_j \delta_j\right)}, 
\end{split}
\end{equation}
where $\alpha_i = 1-\exp{(-{{\sigma}_{obj}}_i \delta_i)}$, and $\delta_i$ is the sampling distance between adjacent points along the ray.
To encourage the rendered 2D object opacity $\hat{O}{(\bm{r})_{obj}^k}$ to satisfy the 2D instance mask, we minimize the squared distance to the corresponding instance mask $M(\bm{r})^{k}$.
We also minimize the squared distance between the rendered object color $\hat{C}{(\bm{r})_{obj}^k}$ and the ground-truth color $C(\bm{r})$ with $M(\bm{r})^{k}$ masked.
The loss of object supervision is defined as:
\begin{equation}
\label{eq:loss_obj}
\begin{split}
    \mathcal{L}_{obj} =  \sum_{\bm{r}\in N_r} \sum_{k\in \llbracket1..K\rrbracket} 
    &\lambda_1 M(\bm{r})^{k} ||\hat{C}{(\bm{r})_{obj}^{k}} - C(\bm{r})||^2_2 \\
     + & \lambda_2 w(\bm{r})^k ||\hat{O}{(\bm{r})_{obj}^{k}}-M(\bm{r})^{k}||^2_2,
\end{split}
\end{equation}
where the instance mask $M(\bm{r})^{k}$ is constructed by setting 1 or 0 w.r.t. the instance label at the corresponding pixel belongs to the object $k$ or not, and $w(\bm{r})^k$ is the balanced weight between 0 and 1 signals of the instance mask.

\noindent\textbf{Occlusion issue.}
The above object supervision is commonly sufficient for learning object radiance field from simple object-centric data (\ie, 360$^{\circ}$ capturing towards a single object without any occlusion).
However, in real-world scene scans, target objects are often occluded by other foregrounds,
which yields incomplete instance masks.
Therefore, for the pixel labels with ``empty'' (or 0) signal at the mask, it is ambiguous whether the ray does not hit the object or is blocked by other foregrounds, and directly using these incomplete masks as supervision may overkill part of the objects and learn a shattered radiance field (see Fig.~\ref{fig:compare_frustum_mask} (c)).
And it's also worth noting that we cannot simply neglect the supervision of the empty areas,
or the model would render unexpected floats at the unsupervised area.

\begin{figure}[t!]
    \centering
    \includegraphics[width=1.0\linewidth, trim={0 0 0 0}, clip]{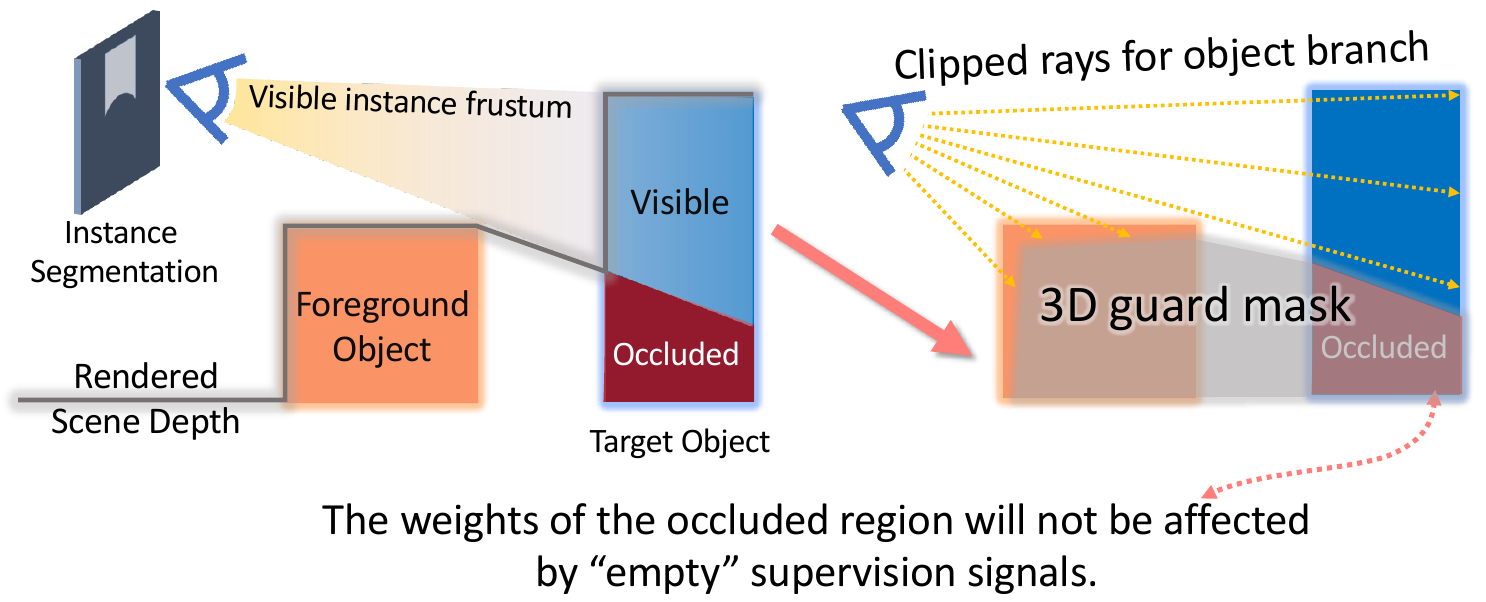}
    \caption{
\textbf{3D guard mask} identifies the occluded region for the object branch.
We render the scene depth $d_{scn}$ (gray lines in the left), and push forward it along the camera direction with $\epsilon$.
Then, we subtract the visible instance frustum (yellow area in the left) from the 3D space farther than $d_{scn}+\epsilon$ to construct the 3D guard mask (gray area).
    }
    \label{fig:3d_guard_mask}
    \vspace{-1.0em}
\end{figure}

\noindent\textbf{Scene-guided occlusion identification.}
We exploit the geometric cues online from the scene branch to identify occlusion regions.
First, we utilize the transmittance from the scene branch to guide the biased sampling of the object branch, which we name scene guidance.
The scene guidance significantly reduces the point sampling inside the occluded region and mitigates the erroneous supervision to the object branch.
However, when the target object is frequently occluded by other instances,
the learned object radiance field is still affected (see Fig.~\ref{fig:compare_frustum_mask} (d)).
Thus, we propose a 3D guard mask to stop the gradient applied to the occluded region, as illustrated in Fig.~\ref{fig:3d_guard_mask}.
In practice, we render the scene depth $d_{scn}$ online by the scene branch
and slightly push forward it along the camera direction with a small distance $\epsilon$.
We then utilize a 3D guard mask to protect the occluded part,
which is constructed by subtracting the visible instance space from the 3D space farther than the pushed scene depth $d_{scn}+\epsilon$.
During the training process of the object branch,
we explicitly prune the ray samples inside the 3D guard mask.
Intuitively, we assume that the distance between two annotated objects is usually larger than $\epsilon$.
Thus, if the target object can be viewed without any occlusion,
our 3D guard mask would permit sufficient point samples for objects and the surrounding space,
so the instance supervision signal will guide the object branch to encode the target object and eliminate everything else.
Otherwise, if the target object is occluded,
the gradient of the ``empty' signal at the occluded region will be blocked,
while the visible region can still be supervised properly.

\subsection{Joint Optimization}

We jointly optimize the scene branch and object branch at the training stage.
For the scene branch,
we follow~\cite{nerf} and minimize the squared error between predicted color $\hat{C}(\bm{r})_{scn}$ and true pixel color $C(\bm{r})$, as:
\vspace{-0.3em}
\begin{equation}
    \mathcal{L}_{scn} =  \sum_{\bm{r}\in N_r} ||\hat{C}(\bm{r})_{scn}-C(\bm{r})||^2_2,
\end{equation}
For the object branch, we use the loss introduced in Eq.~\eqref{eq:loss_obj}.
The total loss of the model is defined as:
\vspace{-0.3em}
\begin{equation}
    \mathcal{L} = \mathcal{L}_{obj} + \mathcal{L}_{scn}.
\end{equation}

\subsection{Editable Scene Rendering}

Thanks to the object-compositional NeRF,
we can readily obtain radiance fields for each annotated object by simply switching the applied optimized object activation code, making it easy to realize the editable scene rendering.
As illustrated in Fig.~\ref{fig:main}, we divide the total editable scene rendering pipeline into the background stage and the object stage.
At the background stage, we obtain the scene color and opacity $\{{\mathbf{c}_{scn}}_i,{\sigma_{scn}}_i\}_{i=1}^N$ from the scene branch while pruning the point sampling at the target region, so as to remove the original objects from the scene.
At the object stage, we shoot rays on the $K$ target objects, and follow the user-defined manipulation to transform the object-specific color and opacity $\{{\mathbf{c}^k_{obj}}_i,{\sigma^k_{obj}}_i\}_{i=1 \; k=1}^{N \;\;\; K}$ to the desired location.
Finally, we aggregate all the opacity and colors by ordering the distance along the ray directions and render pixel colors with the quadrature rules~\cite{quadrature_rule}:
\begin{equation}
\begin{split}
    &{\hat{C}(\bm{r})} = \sum_{i=1}^{N \times (K+1)}
    T_i \alpha_i {\mathbf{c}}_i,
\end{split}
\end{equation}
where $T_i$ and $\alpha_i$ are transmittance and alpha value as defined in Sec.~\ref{ssec:object_learn}.

\section{Experiments}
We evaluate our method in two real-world datasets.
First, we quantitatively and qualitatively compare our scene branch with SoTA methods on standard scene rendering.
Then, we show the visualization of different ways to render the individual object and compare our editable scene rendering with point cloud based rendering method~\cite{npcr}.
Finally, we perform ablation studies to inspect the design of our framework.

\subsection{Dataset}

\noindent\textbf{ToyDesk.}
We created a dataset with instance annotations to evaluate our framework, which contains two sets of posed images with 2D instance segmentation for target objects.
Specifically, we prepare two scenes of a desk by placing several toys with two different layouts, and 360$^{\circ}$ capture images by looking at the the desk center, where the toys are frequently occluded by each others from the image view.
We use the SfM~\cite{colmap}, multi-view stereo~\cite{XuT19} and mesh generation technique~\cite{kazhdan2006poisson} to recover camera poses and meshes, and also manually label the target object on the meshes.
The 2D instance segmentation is obtained by directly projecting the annotated instance labels from the 3D meshes.

\noindent\textbf{ScanNet.}
ScanNet~\cite{scannet} dataset contains RGB-D indoor scene scans as well as 3D instance annotation and 2D instance segmentation by projection.
To better evaluate the performance both of the scene branch and the object branch, we select frames with a viewing angle less than ${40}^\circ$ of a preset central object and distance within 3 meters from the central object, and randomly sample 80\% for training and other frames for testing.
For the object opacity supervision, we directly use the instance segmentation provided in the ScanNet dataset, which is fairly rough (see Fig.~\ref{fig:bonus_seg}) but can be fully leveraged by our method.

Please refer to the supplementary material for more details of the datasets.

\subsection{Data Preparation and Experiment Details}
Our method does not require sensor depth for training as well as NeRF \cite{nerf} and NSVF \cite{nsvf} \footnote{NSVF actually uses depth for the ScanNet dataset.}, while NPCR~\cite{npcr} relies on the depth frames to generate voxel aggregation (according to the author's codebase) and dense point clouds to storage point descriptors.
So for a fair comparison, we follow the training setup in NSVF by adding a depth loss to the training for all the other competitors (\ie, NeRF, NSVF, and our method) on the ScanNet dataset.
However, as the depth loss is totally optional, we will show more results without depth supervision on the ScanNet dataset in the supplementary material.
Besides, as NSVF also utilizes the point clouds for voxel initialization in their ScanNet experiment, we thus use the same point clouds as NPCR to initialize voxels for NSVF and our method.
For the experiments on the ToyDesk dataset, we exclude the NPCR due to the lack of sensor depth and use the SfM point clouds for voxel initialization.

\subsection{Comparison of Scene Rendering}

\begin{figure*}[t!]
    \centering
    \vspace{-1.5em}
    \includegraphics[width=1.0\linewidth, trim={0 0 0 0}, clip]{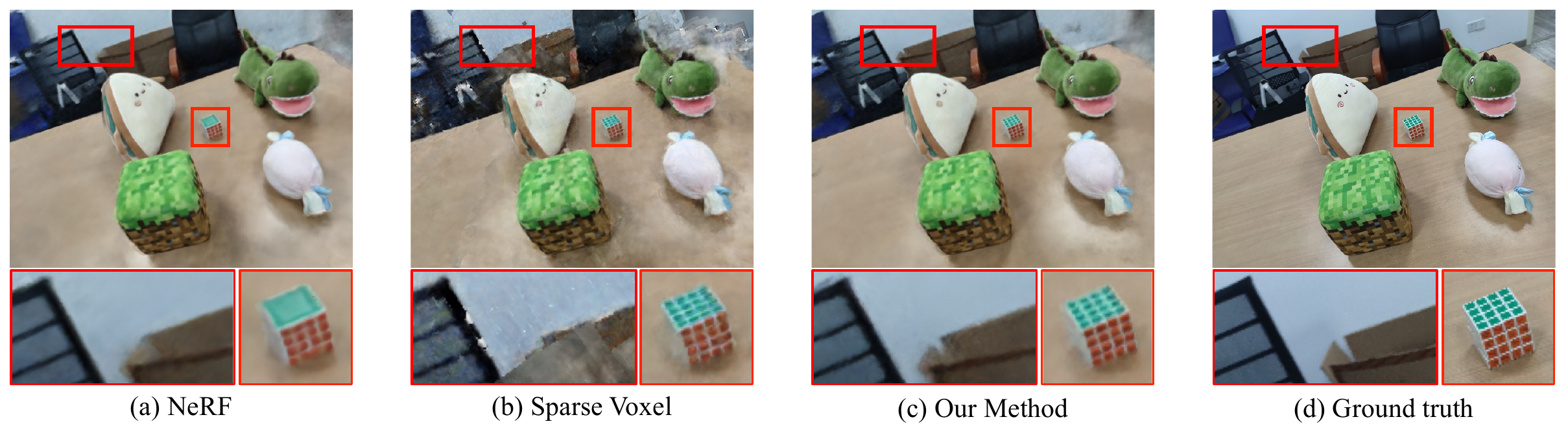}
    \caption{
We compare scene rendering quality with NeRF~\cite{nerf} and Sparse Voxel~\cite{nsvf} on the ToyDesk dataset.
Our method consistently renders fine details of central objects as well as the surrounding environment textures.
    }
    \label{fig:compare_desk}
    \vspace{-1.0em}
\end{figure*}

\begin{figure*}[t!]
    \centering
    \includegraphics[width=1.0\linewidth, trim={0 0 0 0}, clip]{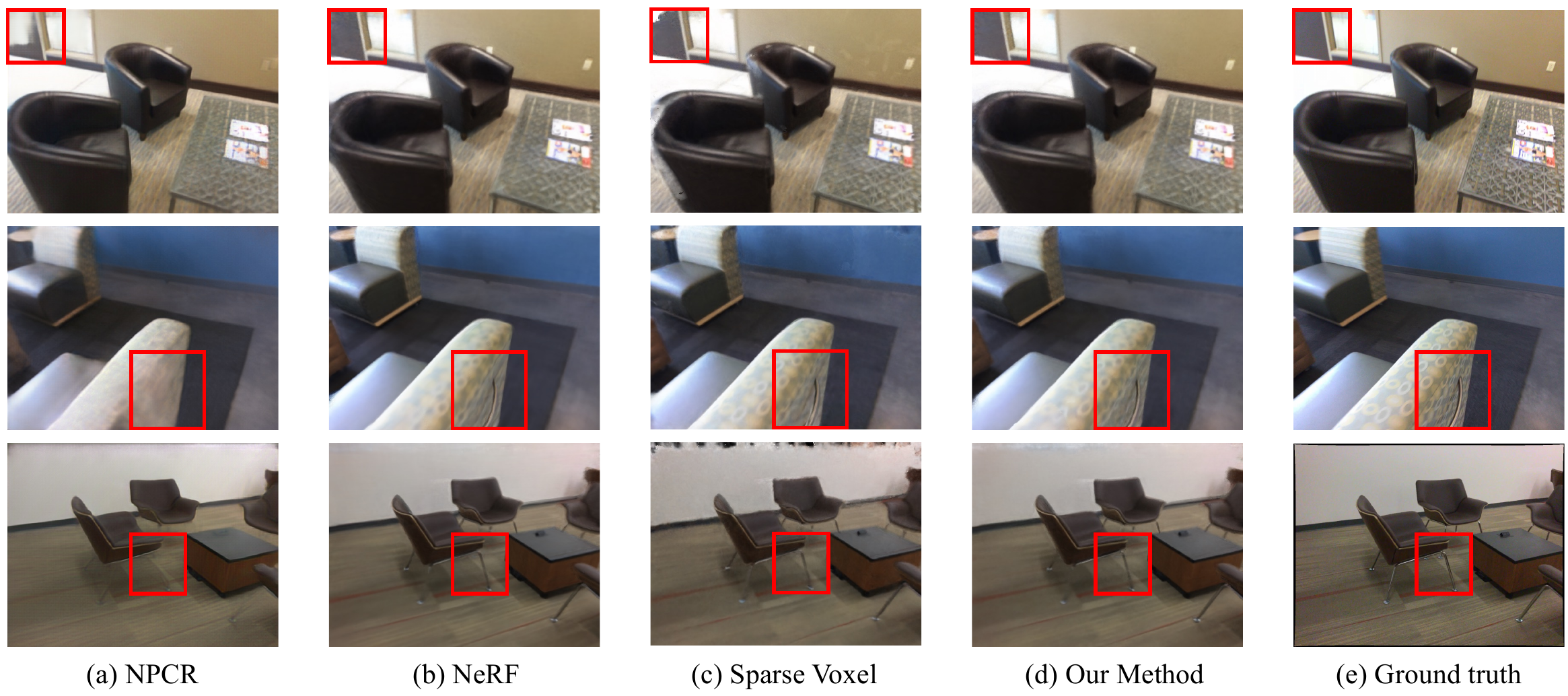}
    \caption{
    We show scene rendering examples of NPCR~\cite{npcr}, NeRF~\cite{nerf}, Sparse Voxel~\cite{nsvf} and our method on the ScanNet dataset.
    Please zoom in for more details.
    }
    \label{fig:compare_scene}
    \vspace{-1.5em}
\end{figure*}

\begin{table}[tb]
\centering
\resizebox{1.0\linewidth}{!}{
\begin{tabular}{lcccccc}
\toprule
\multicolumn{1}{c}{\multirow{2}{*}{Methods}} & \multicolumn{3}{c}{ToyDesk} & \multicolumn{3}{c}{ScanNet} \\ \cmidrule(lr){2-4} \cmidrule(lr){5-7}  
\multicolumn{1}{c}{} & \multicolumn{1}{l}{PSNR $\uparrow$} & \multicolumn{1}{l}{SSIM $\uparrow$} & \multicolumn{1}{l}{LPIPS $\downarrow$} & \multicolumn{1}{l}{PSNR $\uparrow$} & \multicolumn{1}{l}{SSIM $\uparrow$} & \multicolumn{1}{l}{LPIPS $\downarrow$} \\ \hline
NPCR~\cite{npcr} & / & / & / & 25.177 & 0.754 & \textbf{0.225} \\
NeRF~\cite{nerf} & 15.453 & \textbf{0.586} & 0.537 & 28.927 & \textbf{0.815} & 0.249 \\
Sparse Voxel*~\cite{nsvf} & 14.480 & 0.532 & 0.572 & 24.143 & 0.735 & 0.312 \\
Our Method & \textbf{15.607} & 0.585 & \textbf{0.522} & \textbf{29.005} & \textbf{0.815} & 0.243 \\
\bottomrule
\end{tabular}
}
\caption{
To evaluate the full scene rendering quality, 
we quantitatively compare our scene branch with the SoTA neural rendering methods on ToyDesk dataset and ScanNet dataset.
}
\label{tab:scene}
\vspace{-1.0em}
\end{table}

To evaluate the rendering quality of the scene branch, we first compare with neural point cloud based rendering (NPCR)~\cite{npcr}, the state-of-the-art implicit representation method NeRF~\cite{nerf} and the voxel-bounded extension NSVF~\cite{nsvf}.
Since NSVF does not release the training codes for the ScanNet dataset, and we also fail to train the ScanNet data on the official codebase due to the GPU OOM error, we decide to use our implementation of voxel representations along with self-pruning and progressive training mechanism, which will be denoted as Sparse Voxel.

We follow the standard metric in ~\cite{nsvf,nerf,npcr} by using PSNR, SSIM and LPIPS to measure the rendering quality.
As shown in Table~\ref{tab:scene}, our method is comparable or even better than the SoTA methods on the evaluated metrics.
Note that the ToyDesk dataset contains a much larger portion of relatively far background (compared to ScanNet), which degenerates the positional encoding when the query point is far from the origin~\cite{zhang2020nerf++}. Therefore all NeRF-based approaches perform worse on ToyDesk.
Meanwhile, we shows rendered examples for ToyDesk dataset in Fig.~\ref{fig:compare_desk} and ScanNet dataset in Fig.~\ref{fig:compare_scene}.
For the results of ToyDesk, we find that NeRF's outputs look relatively complete, but the details are erased (\eg, the edges of the magic cube), and Sparse Voxel tends to encode more details but fails to produce reasonable texture for far-away backgrounds where the SfM point clouds are incomplete.
Thanks to our \textit{hybrid space embedding} which automatically adapts to varying sampling locations, we consistently show fine details of central objects and the surrounding environment textures.
For the results of ScanNet in Fig.~\ref{fig:compare_scene}, it is noticeable that NPCR and Sparse Voxel both fail to produce the correct color at the first row, and NPCR even omitted the chair handle and the legs as shown in the second and the third row in Fig.~\ref{fig:compare_scene}.
We consider it is mainly due to the incomplete point cloud, which makes it impossible for NPCR and Sparse Voxel to store descriptors in these areas without supporting geometry. 
On the contrary, our method both shows finer granularity where there are voxels and also correctly renders textures where voxel is missing.

\subsection{Comparison of Individual Object Rendering}

We argue that it is unnatural to render the individual object once the entire scene has been encoded to the model, even though the reconstructed 3D mesh or bounding box is available and can be acted as a 3D rendering mask.
To prove this, we compare different approaches to render the individual object in Fig.~\ref{fig:compare_render_obj}, where (b) and (c) are rendered with points sampled close to the mesh surface, with radius 0.05m and 0.1m, respectively, and (e) is rendered with rays clipped by the bounding box,
and (f) is the result from the object branch.
Since the reconstructed mesh is usually incomplete and inaccurate as shown in  Fig.~\ref{fig:compare_render_obj} (a), directly applying a strict 3D rendering mask (\ie within a radius of 0.05m) would produce a mottled rendering result as Fig.~\ref{fig:compare_render_obj} (b) due to the lack of precise point sampling on the main color components of the radiance field.
When relaxing the radius to 0.1m, though the rendered object becomes more vivid and realistic, the background texture is incidentally included as shown in Fig.~\ref{fig:compare_render_obj} (c).
Besides, restricting ray samples inside the bounding box would even include more background textures, as shown in Fig.~\ref{fig:compare_render_obj} (e).
In contrast, our object branch can render a clean object without background included, as shown in Fig.~\ref{fig:compare_render_obj} (f).

\subsection{Comparison of Scene Editing}

\begin{figure}[t!]
    \centering
    \includegraphics[width=0.9\linewidth, trim={0 0 0 0}, clip]{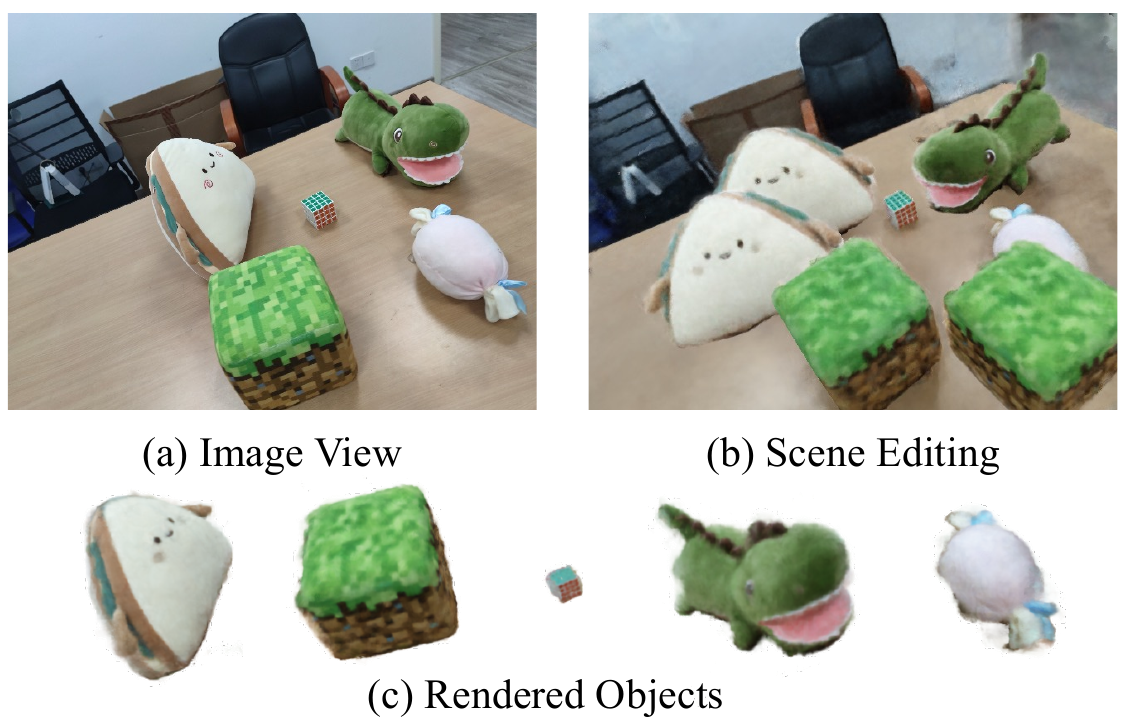}
    \caption{
We show the original image view (a), scene editing result (b) by duplicating and moving toys on the ToyDesk dataset,
and the decomposite objects (b) rendered by the object branch.
    }
    \label{fig:edit_desk}
    \vspace{-1.0em}
\end{figure}

We first show our object rendering and scene editing results on the ToyDesk dataset in Fig.~\ref{fig:edit_desk}.
As shown in Fig.~\ref{fig:edit_desk} (c), the rendered objects from the object branch vividly exhibit the objects with sharp boundaries, which demonstrates the effectiveness of our object-compositional design.
We perform scene editing by rotating, moving, and duplicating objects following the proposed editable rendering pipeline.
From Fig.~\ref{fig:edit_desk} (b), we can see that the manipulated objects are seamlessly integrated into the scene while ensuring the correct spatial relationship.

\begin{figure}[t!]
    \centering
    % \vspace{-1.5em}
    \includegraphics[width=1.0\linewidth, trim={0 0 0 0}, clip]{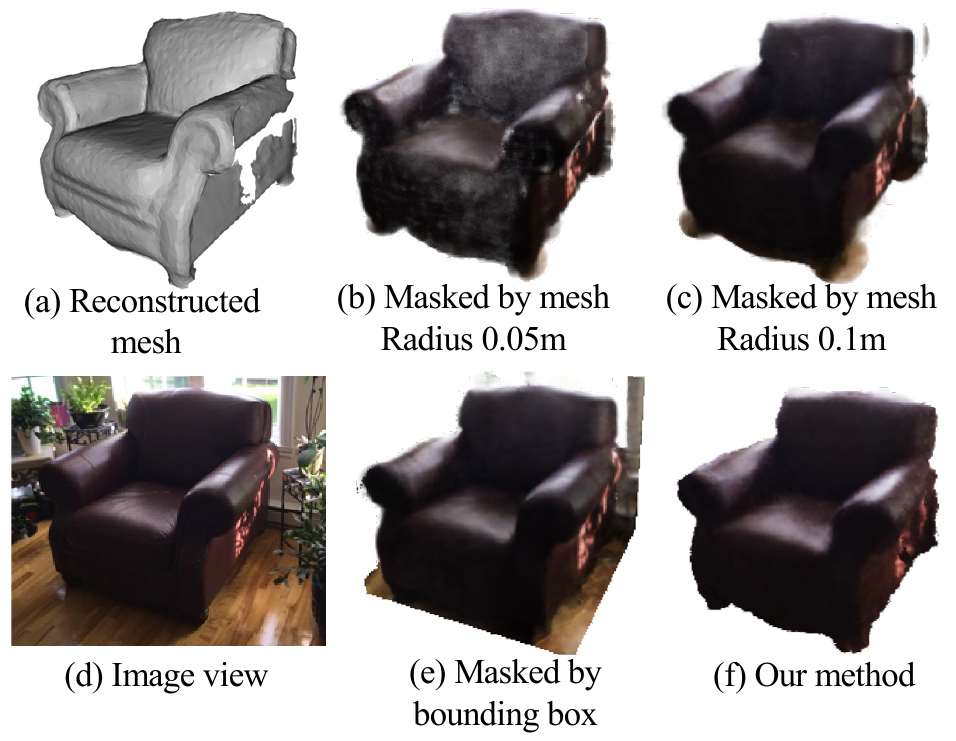}
    \caption{
    We compare different methods to render the individual object on the ScanNet dataset.
    }
    \label{fig:compare_render_obj}
    \vspace{-1.0em}
\end{figure}

\begin{figure*}[t!]
    \vspace{-2.5em}
    \centering
    \includegraphics[width=1.0\linewidth, trim={0 0 0 0}, clip]{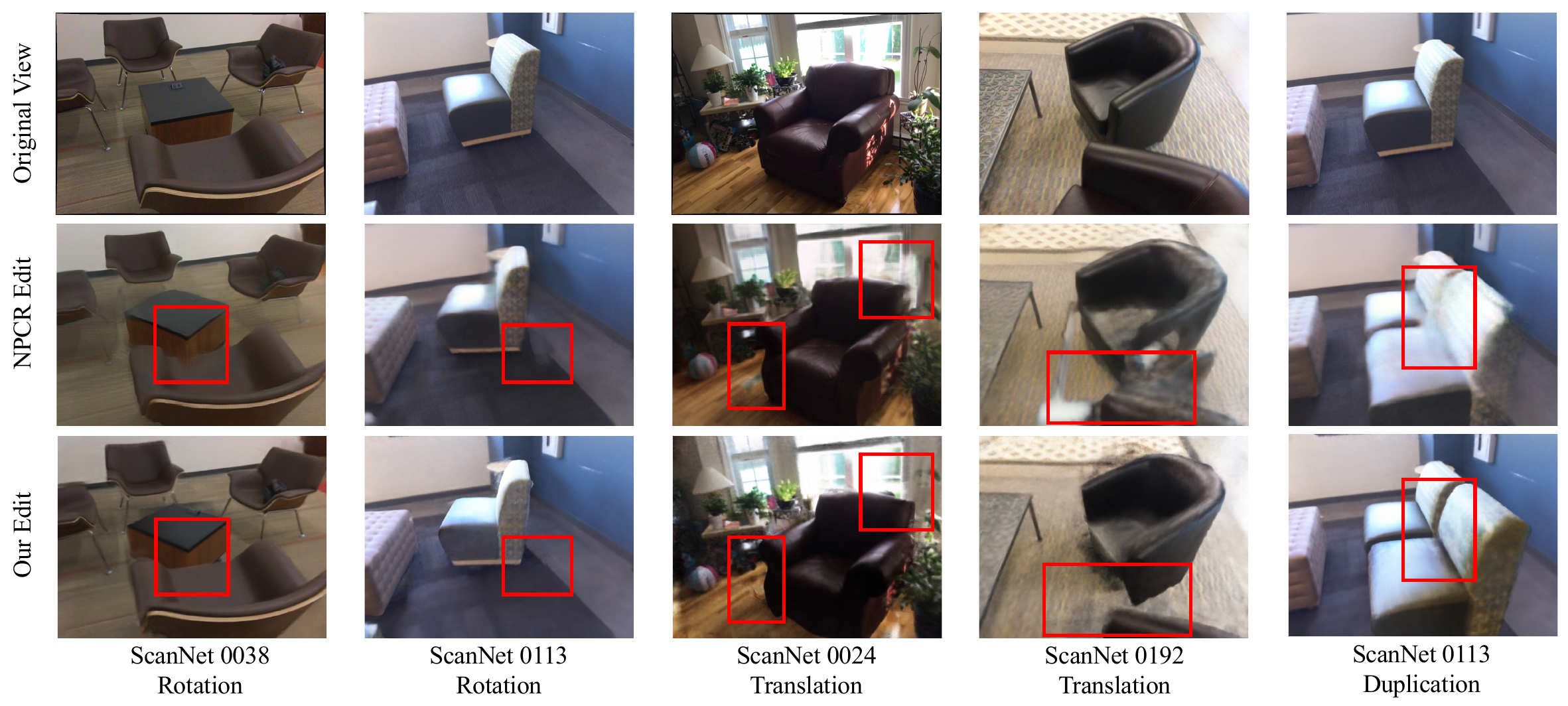}
    \caption{
We compare our method and NPCR~\cite{npcr} for scene editing by rotating, translating, and duplicating objects on the ScanNet dataset.
    }
    \label{fig:compare_edit}
    \vspace{-0.6em}
\end{figure*}

We then compare our editable scene rendering with neural point cloud based rendering method NPCR~\cite{npcr} on the ScanNet dataset.
The NPCR takes the input of raw point clouds of a scene and outputs images of novel views.
During training, NPCR also optimizes feature vectors for each 3D point which encodes scene appearance.
Therefore, we perform scene editing by manually moving point clouds and the feature vectors inside the bounding box of the target objects.
As shown in Fig.~\ref{fig:compare_edit}, we find that the manipulation of point clouds inside the bounding box also brings the movement of background texture (\eg, the pattern of the windows and the carpet under the chair have been moved in `0024 Translation' and `0113 Rotation'), which may be due to NPCR incidentally stores part of the background appearance feature to the points inside the bounding box.
In contrast, our method moves the objects while keeping the textures of the background nearly unchanged,
which produces more realistic editing results.
Besides, NPCR also renders the images with severe artifacts in some cases (\eg, false occlusion relationship in `0038 Rotation', unexpected holes of chairs in `0192 Translation', and cloudy textures in `0113 Duplication').
We guess that it somehow encodes the visibility of 3D points with an implicitly fixed order rather than inferencing from the 3D space, and the aggregation from the mixture of object points and invisible noisy points also confuses the neural renderer.
By the way, we also test scene editing with Sparse Voxel, but find the issue similar to the bounding box based approaches (Fig.~\ref{fig:compare_render_obj}).
However, as we output the radiance field of each target objects independently and the rendering pipeline also takes the advantages of volumetric rendering, we consistently produce realistic editing results with correct space relationship and intact textures.
Please refer to the supplementary material for more evaluation of our editing results.

\begin{figure}[t]
    \centering
    \includegraphics[width=1.0\linewidth, trim={0 0 0 0}, clip]{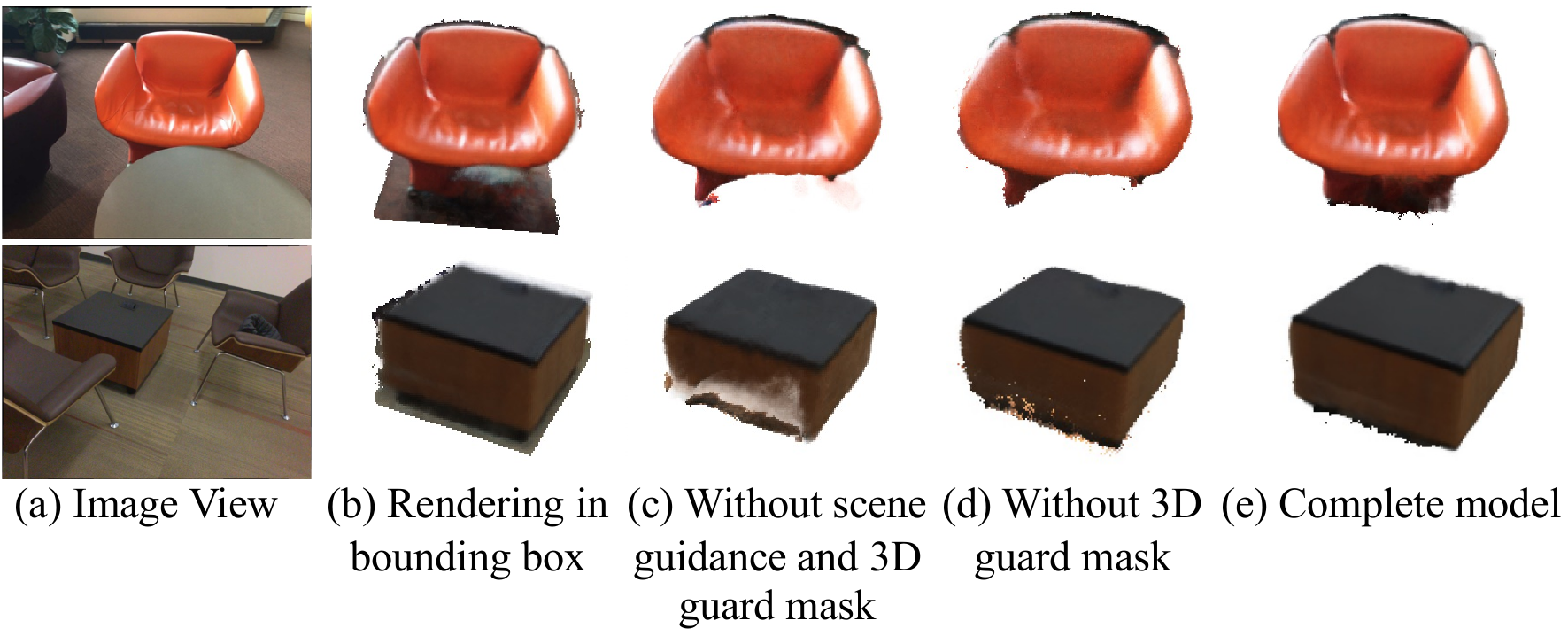}
    \caption{
    We visualize the effectiveness of scene guidance and 3D guard mask by ablating them on the training process of the ScanNet dataset. Note that (b) is produced by the scene branch with sampling rays clipped inside the bounding box and can be considered as a reference complete view of the target object.
    Our scene guidance and 3D guard mask effectively prevent the overkill of the occluded region and ensure a complete object rendering. 
}
    \label{fig:compare_frustum_mask}
    \vspace{-0.9em}
\end{figure}

\subsection{Ablation Study}

\begin{table}[tb]
\resizebox{1\linewidth}{!}{
\begin{tabular}{lcccccc}
\toprule
\multicolumn{1}{c}{\multirow{2}{*}{Config.}} & \multicolumn{3}{c}{ScanNet 0033} & \multicolumn{3}{c}{ScanNet 0038} \\ \cmidrule(lr){2-4} \cmidrule(lr){5-7} 
\multicolumn{1}{c}{} & PSNR $\uparrow$ & SSIM $\uparrow$ & LPIPS $\downarrow$ & PSNR $\uparrow$ & SSIM $\uparrow$ & LPIPS $\downarrow$ \\ \midrule
w/o SG, 3DGM & 19.785 & 0.750 & 0.111 & 27.100 & 0.808 & 0.112 \\
w/o 3DGM & 20.450 & 0.754 & 0.117 & 33.914 & 0.884 & 0.056 \\
w/o $f_{obj}$ & 22.219 & 0.817 & 0.057 & 33.861 & 0.892 & 0.058 \\ 
Complete & \textbf{22.600} & \textbf{0.822} & \textbf{0.049} & \textbf{34.435} & \textbf{0.897} & \textbf{0.056} \\
\bottomrule
\end{tabular}
}
\newline
\caption{Ablation for the effectiveness of our proposed scene guidance, 3D guard mask and object voxel feature on learning object radiance field.}
\label{tab:ablation_main}
\vspace{-1.0em}
\end{table}

\noindent\textbf{Scene guidance and 3D guard mask.}
We analyze the effectiveness of scene guidance and 3D guard mask for learning object-compositional rendering when the target objects are frequently viewed with partial occlusion.
Specifically, we choose two scenes (ScanNet 0033 and 0038) where the target objects are frequently occluded by foreground furniture.
We first randomly select ten testing views for each scene and quantitatively inspect these strategies on the rendered objects in Table~\ref{tab:ablation_main}, where SG denotes the scene guidance (biased sampling distribution) provided by the scene branch, and 3DGM denotes the 3D guard mask.
Practically, to block the influence of the background color during the evaluation, we use the instance segmentation to mask out the background and crop the ground truth and the rendered images to tightly fit the object area.
The results in Table~\ref{tab:ablation_main} show that our scene guidance and 3D guard mask significantly improve the rendering quality of the target objects.
Besides, we also show a visual comparison in Fig.~\ref{fig:compare_frustum_mask}.
Thanks to these strategies, we can learn an intact object radiance field even the target object is rarely observed completely.
More qualitative and quantitative results can be found in the supplementary material.

\begin{table}[tb]
\resizebox{1\linewidth}{!}{
\begin{tabular}{lrrrrrr}
\toprule
\multicolumn{1}{c}{\multirow{2}{*}{Config.}} & \multicolumn{3}{c}{ScanNet 0033} & \multicolumn{3}{c}{ScanNet 0038} \\ \cmidrule(lr){2-4} \cmidrule(lr){5-7} 
\multicolumn{1}{c}{} & \multicolumn{1}{l}{PSNR $\uparrow$} & \multicolumn{1}{l}{SSIM $\uparrow$} & \multicolumn{1}{l}{LPIPS $\downarrow$} & \multicolumn{1}{l}{PSNR $\uparrow$} & \multicolumn{1}{l}{SSIM $\uparrow$} & \multicolumn{1}{l}{LPIPS $\downarrow$} \\ \midrule
$\epsilon$ = 0.025 & 22.364 & 0.819 & 0.051 & 34.047 & 0.895 & 0.057 \\
$\epsilon$ = 0.05 & \textbf{22.600} & \textbf{0.822} & \textbf{0.049} & \textbf{34.435} & \textbf{0.897} & \textbf{0.056} \\
$\epsilon$ = 0.1 & 22.172 & 0.818 & 0.053 & 34.299 & 0.896 & \textbf{0.056}\\
\bottomrule
\end{tabular}
}
\newline
\caption{Quantitative evaluation w.r.t forward distance $\epsilon$ of 3D guard mask on learning object radiance field.}
\label{tab:ablation_mask_eps}
\vspace{-0.7em}
\end{table}

\noindent\textbf{Choice of different $\epsilon$ in 3D guard mask.}
To study the impact of different $\epsilon$ in 3D guard mask, we vary the $\epsilon$ and report the metric evaluation of the object area as introduced above. From Table~\ref{tab:ablation_mask_eps}, we find that $\epsilon=0.05$ achieves better rendering quality.
As the other choices also produce very close results, we believe our proposed 3D guard mask is not sensitive to the choice of $\epsilon$.

\noindent\textbf{Object voxel feature.}
To inspect the effectiveness of object voxel feature $\bm{f}_{obj}$ illustrated in~Fig.~\ref{fig:framework},
we ablate it by removing the embedded $\bm{f}_{obj}$ at the input of the object branch and quantitatively evaluate the object area as we introduced above.
As shown in Table~\ref{tab:ablation_main}, the design of $\bm{f}_{obj}$ further boosts the rendering quality of the objects, which indicates adding learnable parameters in 3D space can also broaden the network ability for compositional rendering.

\begin{figure}[t]
    \centering
    \includegraphics[width=0.97\linewidth, trim={0 0 0 0}, clip]{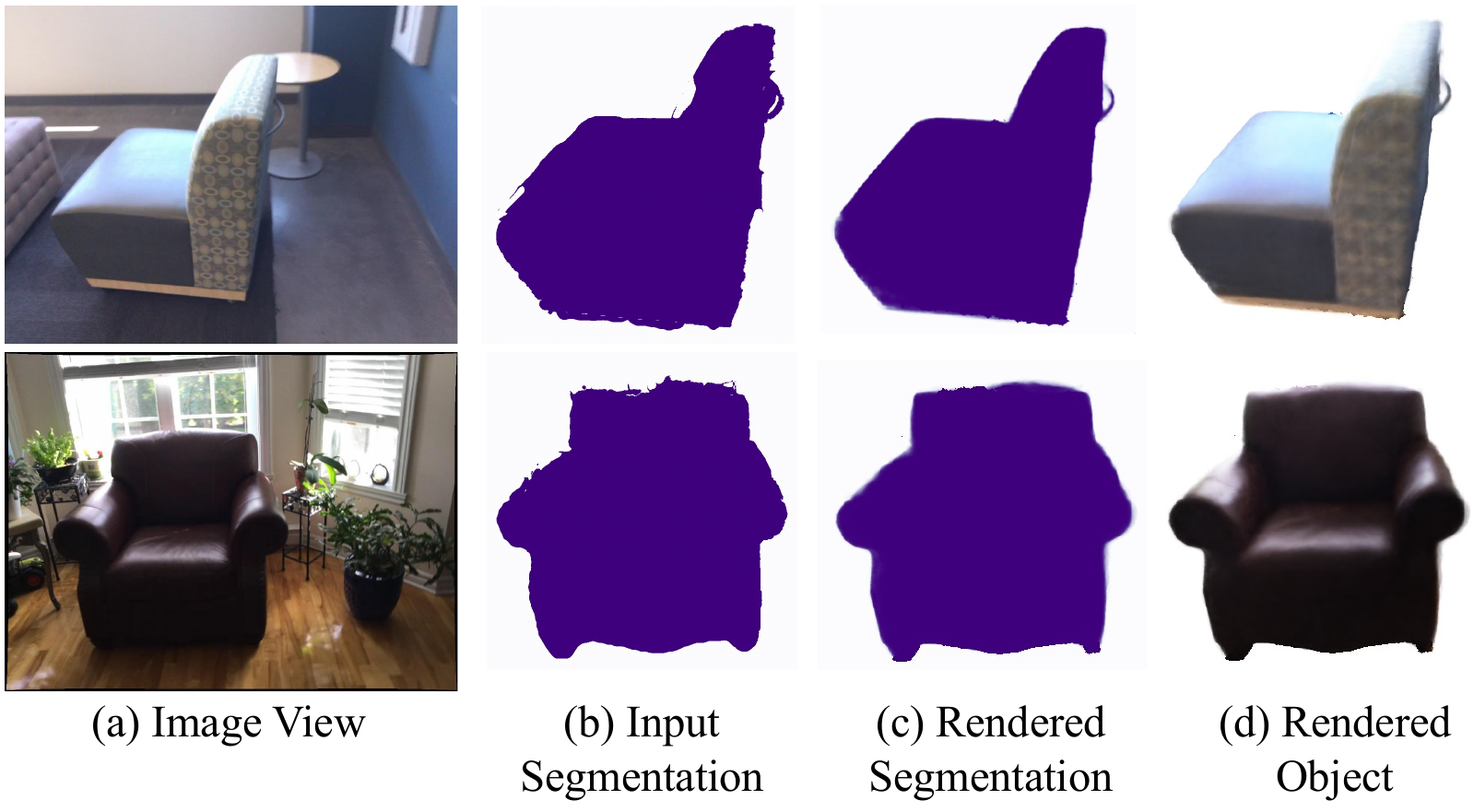}
    \caption{
    We show the image view (a),
    input segmentation for supervision (b), our rendered segmentation (c), and the rendered object (d).
    Please zoom in for more details.
    }
    \label{fig:bonus_seg}
\end{figure}

\subsection{Visualization of Rendered Segmentation}

Since our method only relies on the 2D segmentation to learn the decomposition of the target objects, we visualize the input segmentation used for supervision and our rendered segmentation (2D object opacity) as well as the rendered objects in Fig.~\ref{fig:bonus_seg}.
To our surprise,  even though the input segmentation is fairly rough with jittery edges, our method can produce a smooth and accurate segmentation once the training converges while preserving high-fidelity details of the object (\eg, chair handle at the first row).
We believe the multi-view supervision helps resist mask noise from a single view, and the converged 3D structure learned from images provides geometry cues for object decomposition, which is also observed in a parallel work by Zhi \etal~\cite{zhi2021place}.
This shed light on distilling fined-grained 3D segmentation only from the knowledge of 2D segmentation networks by the proposed learning pipeline.

\section{Conclusion and Future Works}
We present the first neural scene rendering framework which provides high-fidelity novel-view synthesis while supporting editable scene rendering on real-world scenes.
By training with the posed images and rough 2D instance masks, we can freely utilize the model to render novel views with multiple objects manipulated (\eg moving, rotating, or duplicating).
Currently, due to the lack of observations, our method relies on the network spatial smoothness to render unseen textures under objects, which can be further optimized by adopting scene completion methods.
To mitigate the influence of the pose noise and rolling shutter on the real-world data, we can further optimize camera poses and ray directions so as to render a clearer background.
Besides, to achieve more realistic scene editing, it is also promising to integrate the scene lighting model into the framework in the future work.

\noindent\textbf{Acknowledgments:}
We thank Hanqing Jiang, Liyang Zhou and Jiaming Sun for their kind help in scene reconstruction and annotation for the ToyDesk dataset. This work was partially supported by NSF of China (No. 61932003).

\clearpage

{\small
\bibliographystyle{ieee_fullname}
\bibliography{main}
}

\end{document}